\documentclass[sn-mathphys]{sn-jnl}

\jyear{2024}%

\theoremstyle{thmstyleone}%
\theoremstyle{thmstyletwo}%

\theoremstyle{thmstylethree}%
\usepackage{dsfont}

\begin{document}

\title[Geometry Preserving Loss Functions Promote Improved Adaptation of Blackbox Generative Models]{Geometry Preserving Loss Functions Promote Improved Adaptation of Blackbox Generative Models}


\author[1]{\fnm{Sinjini} \sur{Mitra}}
\equalcont{Corresponding Author}

\author[1]{\fnm{Constantine} \sur{Kyriakakis}}


\author[2]{\fnm{Shenyuan} \sur{Liang}}

\author[3]{\fnm{Eun Som} \sur{Jeon}}
\equalcont{Corresponding Author}

\author[4]{\fnm{Rushil} \sur{Anirudh}}

\author[5]{\fnm{Jayaraman J.} \sur{Thiagarajan}}

\author[2]{\fnm{Anuj} \sur{Srivastava}}

\author[1]{\fnm{Pavan} \sur{Turaga}}


\affil[1]{\orgdiv{Geometric Media Lab}, \orgname{Arizona State University}, \orgaddress{\city{Tempe} \postcode{85281},  \country{USA}}}

\affil[2]{\orgdiv{Department of Statistics}, \orgname{Florida State University}, \orgaddress{\city{Tallahassee} \postcode{32306},  \country{USA}}}

\affil[3]{\orgdiv{Department of Computer Science and Engineering}, \orgname{Seoul National University of Science and Technology}, \orgaddress{\city{Seoul} \postcode{01811},  \country{Republic of Korea}}}

\affil[4]{\orgname{Amazon}, \orgaddress{\country{USA}}}

\affil[5]{\orgname{Apple Inc.}, \orgaddress{\country{USA}}}


\abstract{Adaptation of blackbox generative models has been widely studied recently through the exploration of several methods including generator fine-tuning, latent space searches, leveraging singular value decomposition, and so on. However, adapting large-scale generative AI tools to specific use cases continues to be challenging, as many of these industry-grade models are not made widely available. The traditional approach of fine-tuning certain layers of a generative network is not feasible due to the expense of storing and fine-tuning generative models, as well as the restricted access to weights and gradients. Recognizing these challenges, we propose a novel end-to-end pipeline aimed at domain adaptation by leveraging geometry-preserving loss functions in conjunction to pre-trained generative adversarial networks (GANs). Our method rethinks the problem of adaptation by re-contextualizing the role of GAN inversion in obtaining accurate latent space representations. Extending the ability of existing state-of-the-art inverters, we preserve pair-wise distances between tangent spaces to successfully train a latent generative model to produce samples from the target distribution. We evaluate our proposed pipeline on StyleGANs with real distribution shifts and demonstrate that the introduction of the geometry preserving loss function lends to improved adaptation of generative models compared to other traditional loss functions.}

\keywords{Domain adaptation, manifold learning, latent space generative model, geometric prior.}



\maketitle

\section{Introduction}\label{intro}

In recent years, generative networks have shaped the landscape of the machine learning community.  The recent successes in this field have brought to light certain challenges in adapting such large-scale generative models \cite{kwon2023one, yang2023one, gan2023decorate, zhang2022transfer, zhumind}  due to their size and dependence on large amount of target domain data. Some recent methods leverage CLIP-guided latent optimization with contrastive regularization \cite{kwon2023one}, test-time adaptation through prompt learning \cite{Boudiaf_2022_CVPR}, miner based approaches to find the correct latent manifolds \cite{wang2024minegan++}, training auxiliary networks coupled with GAN retraining, etc. to subvert this assumption. Although these methods report an improved adaptation performance, there still exists a need for either generator fine-tuning or the introduction of an additional network; or both, to perform latent optimization. These de facto solutions are often undesirable due to the exorbitantly large memory footprints, data scarcity in custom domains, and limited compute availability in practical scenarios.

From a deployment standpoint, important potential ethical and legal implications and responsibilities are associated with making industry-grade model weights publicly available. With recent advances in generative models, there is a renewed focus and debate on the far-reaching consequences of making existing models available to the public \cite{floridi2023ethics, schlagwein2023chatgpt,rane2023chatgpt}.  To address these concerns, many models are only accessible through an API call in a black-box fashion (such as DALL-E \cite{dalle} and Midjourney \cite{Midjourney}). However, there is a noticeable increase in the presence of bad-faith actors and exploitation of AI for non-productive applications. Thus, there is a strong need for models and pipelines that are able to function while addressing ethical and legal concerns for the general public. Particularly, in the case of generative models, it is not clear how model adaptation can be realized in such a setting where the weights are not accessible but there is a need to sample from a target distribution.


There is increasing interest in leveraging the geometric properties of latent-spaces in areas like domain adaptation \cite{thopalli2023surprising, wu2020geometric}, where the idea is that imposing simplified geometries helps in faster adaption, with small training sets. However, prior attempts have made very restrictive assumptions on geometric structures such as subspaces or collections of subspaces. There is not much work in using such constraints for the task of generative model adaptation, partly because in large-scale generative models, the latent spaces are more complex than can be modeled by simple structures like subspaces. This requires more expressive ways to encode latent spaces geometries. There is a long body of work on considering the set of images that result from various geometric and photometric variations in scenes, and express the variation in terms of low-dimensional manifolds, embedded in high-dimensional image-space \cite{Nayar_CAVE_0213,Pless2009}. Manifold learning has been revisited with deep-learning models, including generative models \cite{Ni2022_GANs,Shamsolmoali2023}. However, there is no well-accepted approach to impose a `manifold-prior' as a simple loss function. 

In this paper, we take steps to enforce basic manifold properties explicitly in distances and tangent planes, in both the image and latent space, which are encoded expressly in the adaptation objective, and we show that this helps in preserving the geometric aspects leading to improved adaptation.
To advance past the existing limitations, we propose a robust method for adaptation as an alternative to fine-tuning. We leverage the manifold prior offered by introducing a geometry preserving loss term and achieve adaptation through three separate and independent steps: (1) inverting target domain images to the latent space of a source generative model, (2) learning a latent sampler to directly sample from the sub-manifold corresponding to the target images, and (3) converting the sampled target latents to images. We explore the compatibility of our pipeline with various existing inversion methods and design the latent sampler in the 18$\times$ 512 space as a 1-D diffusion model. The inversion is performed using state-of-the art inverters and the sampled latents are translated to images using the source generator, both of which are kept frozen. Our pipeline improves adaptation performance and makes the following contributions:

\begin{enumerate}
\item Preserving the geometry in the image and latent space enforces a `manifold-prior' that promotes improved adaptation for OOD images.
\item We completely remove the need for access to the weights of large-scale pre-trained models thus realizing adaptation in a more secure way.
\item Our method is robust to limitations on the availability of target data and is able to achieve good results for a small number of available target images.
\end{enumerate}

\section{Related Work}
\label{sec: related}

GAN adaptation \cite{9802910, zhang2022generalized, bejiga2018gan, sanabria2021unsupervised} is the problem of approximating a target distribution $P_{\text{target}}(X)$ given access to  (a small number of) target samples, $\{\mathrm{X}_i\sim P_{\text{target}}\}$ using a generator that is pre-trained on a training distribution, $G(\mathrm{z}) \sim P_{\text{base}}$, where $\mathrm{z}\sim P(\mathrm{z})$ is a known prior. The most straight-forward approach for adaptation is GAN fine-tuning \cite{pmlr-v119-zhao20a, mo2020freeze, karras2019style, sauer2021projected}, which is impractical in reality. Fine-tuning generative models is not only compute-heavy, but it also assumes that a substantial amount of target data is widely available. One shot adaptation strategies \cite{chong2022jojogan, thopalli2023sista} leverage clever data augmentation tricks, while other methods leverage existing models such as CLIP \cite{radford2021learning} to bypass this assumption \cite{kwon2023one, zhumind, zhang2022towards}. Certain approaches adapt a pre-trained GAN by finding the singular value decomposition of the generator and discriminator weights (c.f. \cite{robb2020few}). Many other approaches introduce auxiliary networks to help retrain parts of a pre-trained GAN to adapt to a novel target domain \cite{yang2023one, wang2024minegan++}.

The recent advances made in diffusion models \cite{ho2020denoising, song2020denoising} have brought to light their excellent capability in learning probability distributions and hence caused a paradigm shift in the generative modeling landscape. In the context of domain adaptation, text-to-image diffusion models have shown considerable promise in the one and few-shot setting for both 2D and 3D generation \cite{Benigmim_2023_CVPR, Kim_2023_CVPR, Song_2024_WACV}. CLIP-guided diffusion models have also applied the CLIP objective to diffusion generators \cite{kim2022diffusionclip}. Other methods fine-tune the text embedding \cite{gal2022image}, or the entire diffusion model  based on a few target images \cite{ruiz2023dreambooth}. A recent work \cite{Li_2024_CVPR} proposes aligning the StyleGAN $W^{+}$ space with the stable diffusion model using a mapping network and a residual cross-attention module. The model is guided to produce personalized images through text descriptions. 

In existing works that leverage geometric priors for adaptation \cite{thopalli2023surprising, wu2020geometric, zhan2019ga} much of the focus has been on achieving efficient cross-domain shifts. A geometry-aware GAN was proposed by Fu {\em et. al} that preserves cyclic reconstructions and distance between pairs of images from different domains \cite{Fu_2019_CVPR}. By carefully balancing geometry preservation and the original GAN objective, the model by geometry-aware GAN reports marked improvement in one-side unsupervised domain adaptation. However, this approach is not suitable for a blackbox setup where modifying the source generator (or discriminator) is not possible. Thus, existing methods rely on the availability of large-scale models for parameter optimization, generator editing, and model retraining. However, novel target domains may not have the large amounts of data needed to retrain models or edit existing generator weights. 

Considering these limitations, we propose adaptation by inverting the target dataset onto the latent space of the base generative model ($G_s$), and subsequently training a target sampler with the help of a geometric loss that aids in estimating the latent (target) distribution. This formulation is agnostic to the choice of inversion technique and can be achieved through any pre-trained encoder (e.g., pSp\cite{richardson2021encoding} and e4e \cite{tov2021designing}) or the target sampler design. It is important to note that our pipeline is not dependent on text-based prompts or generator fine-tuning to guide the adaptation. 

\section{Preliminaries}
\label{sec: method}

\paragraph{Problem formulation.}

\begin{figure*}[ht]
    \centering
    \includegraphics[width=\textwidth]{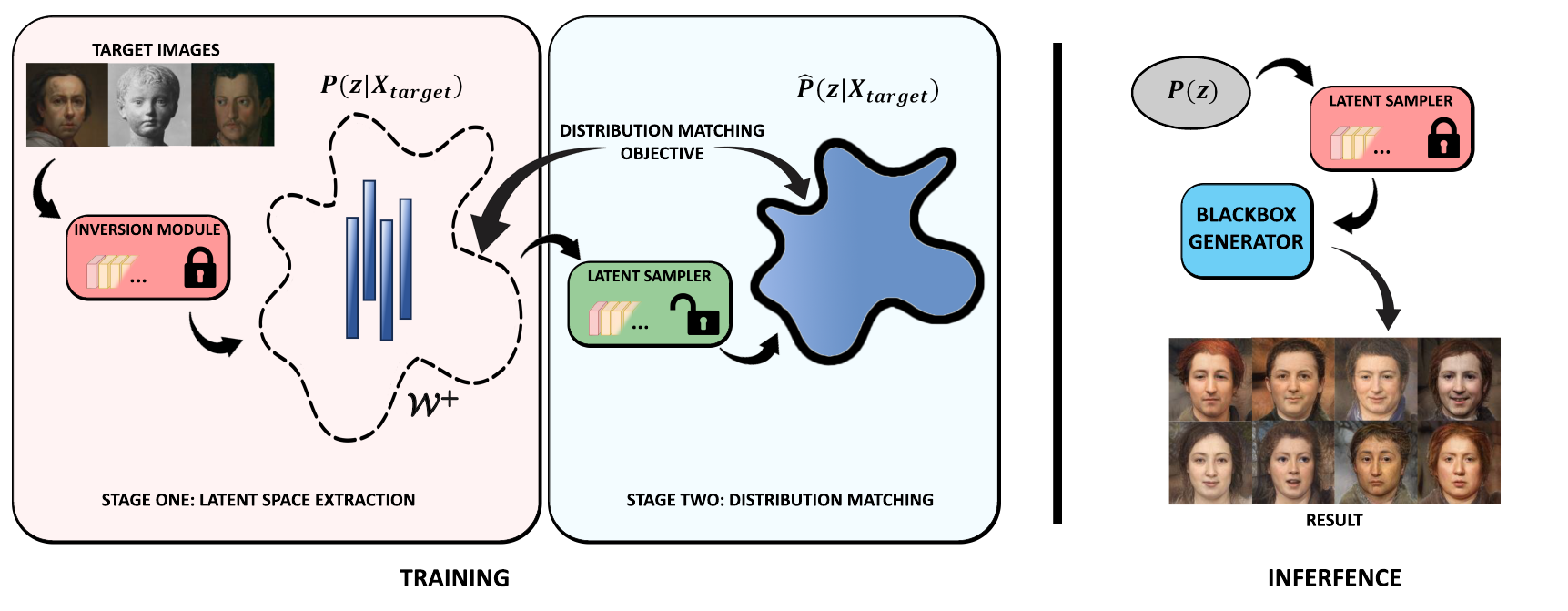}
    \caption{Overview of our pipeline consisting of three stages - Stage 1: Given a set of target images, the corresponding set of target latents is obtained using the frozen inversion module, Stage 2: The inverted latents are input to the latent sampler which is trained with a distribution matching objective that enables it to learn to sample from the estimated posterior, Stage 3: During inference, the trained target sampler uses a gaussian prior to synthesize realizations from the learned target distribution to generate target domain images.}
    \label{fig:overview}
\end{figure*}

Given the small number of target samples $\{\mathrm{X}_i\sim P_{\text{target}}\}$, we use a pre-trained inverter to obtain the corresponding latents $\mathrm{z}_{\text{target}}$ by projecting the images to the sub-manifold of the source generator ($G_s$) to obtain $P(\mathrm{z} \mid \mathrm{X}_{\text{target}})$. Then, using a generative model in the latent space, we map a known prior distribution $P(\mathrm{z})$ to match the target latents obtained via inversion. Figure \ref{fig:overview} gives an overview of our method. In the following sections, we address related concepts and modules for our pipeline.

\paragraph{StyleGAN2.}

The StyleGAN framework \cite{karras2019style} consists of two parts: (1) a mapping network that transforms the noise $(\mathrm{z})$ sampled from a distribution $P(\mathrm{z})$ in $\mathcal{Z}=\{ \mathrm{z} \mid P(\mathrm{z}) \neq 0\} \subseteq \mathbb{R}^{1 \times 512}$ to the latent space $\mathcal{W} \subseteq \mathbb{R}^{1 \times 512}$ using a fully connected network, $F$, and (2) a synthesis network $G$ that translates the latent codes from $\mathcal{W}$ to the image space $\mathcal{X} = \{G(\mathrm{w}) \mid \mathrm{w} \in \mathcal{W}\} \subseteq \mathbb{R}^{3\times H \times W}$.  In our work, the latent sampler learns to estimate $P(\mathrm{z}\mid
\mathrm{X}_{\text{target}})$  which gives rise to latents and not images. Therefore, we use the pre-trained StyleGAN2 generator, $G_s$ in inference mode to map the generated  latent codes to the image space ($\mathcal{X}$). It is important to note that when generating images, $G_s$ is kept frozen and thus no update is made to its weights.

\paragraph{GAN inversion.}

GANs contain two modules - a generator $G$ and a discriminator $D$ that partake in an adversarial game to outperform each other. The generator learns the mapping  between the latent space $\mathcal{Z}$ and image space $\mathcal{X}$, $G : \mathcal{Z}  \rightarrow \mathcal{X}$. GAN inversion \cite{zhu2016generative,xia2022gan, wang2022high} is the reverse of the mapping above, i.e., from $\mathcal{X} \rightarrow \mathcal{Z}$ where an image $\mathrm{x} \sim P(\mathcal{X})$ is mapped to a specific latent code $\mathrm{\textbf{z}}^* \in \mathcal{Z}$ . This can be done by finding an image $\mathrm{x}^*$ that can be entirely synthesized by $G$ such that it is close to the original image $\mathrm{x}$. Formally, if $\mathrm{x} \in \mathbb{R}^n$, $G: \mathbb{R}^k \rightarrow \mathbb{R}^n (k < n)$, and $\mathrm{\textbf{z}} \in \mathbb{R}^k$ (where $k$ is the assumed dimensionality of the latent space), the inversion problem is defined as:
%
\begin{equation}
\mathbf{z}^*=\underset{\mathbf{z}}{\arg \min } \: \ell(G(\mathbf{z}), \mathrm{x}),
\end{equation}
where $\ell$ is a distance metric. Thus, GAN inversion is an unique and powerful tool through which target domain images can be mapped to an approximate sub-manifolds of the latent space of the source generator, $G_s$.   In our specific use case of domain adaptation, this property is extremely useful in navigating adaptation in the instance where source generator fine-tuning is not available. It has been established through prior research that it is difficult to embed directly in the $\mathcal{Z}$ or $\mathcal{W}$ space \cite{abdal2020image2stylegan++} of a StyleGAN2 generator. Thus, we instead use inverters that map directly to the extended latent space $\mathcal{W}^+$, which is a concatenation of 18 different 512-dimensional $\mathrm{w}$ vectors. The $\mathcal{W}^+$ space provides better reconstruction performance of inverted images and is therefore the right choice for our pipeline.

\paragraph{Latent space sampler.}

The design of the latent sampler can be achieved through numerous architecture choices (such as miners \cite{wang2024minegan++} and normalizing flows\cite{papamakarios2021normalizing}). However, due to the excellent ability of recent diffusion models (such as DALL-E and Midjourney) to estimate probability distributions, in this paper, we take inspiration from denoising diffusion models and choose to design a 1D diffusion model \cite{ho2020denoising} with a 1D U-Net \cite{ronneberger2015u} backbone, $F: \mathbb{R}^{18 \times 512} \rightarrow\mathbb{R}^{18 \times 512}$, for the latent sampler design. The U-Net model is lightweight with only 440K parameters, which is $\sim$ 1.6 \% of
the number of parameters in the base generator ($\sim$ 28M).

Given a set of target images  $\{\mathrm{X}_i\sim P_{\text{target}}\}$, we obtain the corresponding target latents $\mathrm{\textbf{z}}_{\text{target}}$ ($N \times 18 \times 512$, where $N$ is the total number of target images) through the inversion module. These latents represent the distribution  $P(\mathrm{z}\mid \mathrm{X}_{\text{target}})$ which the diffusion model estimates as $\hat{P}(\mathrm{z}\mid\mathrm{X}_{\text{target}})$. The input to the diffusion model are batches of $\mathrm{\textbf{z}}_i \sim \mathrm{\textbf{z}}_{\text{latent}}$ where each $i^{th}$ batch is of size $B \times 18 \times 512$ . In general, diffusion models learn a probability distribution by gradually denoising a random variable distributed according to a known prior \cite{dhariwal2021diffusion}.  The model is parametrized as $\epsilon_{\theta}(\mathrm{z}_t, t)$ where $\mathrm{z}_t = \mathrm{z}_0 + \epsilon(t)$, $0<t<T$ and $\epsilon(t)$ is additive noise at timestep $t$. Through a noising-denoising process, the diffusion model produces an estimate of the input latents,  $\hat{\mathrm{\textbf{z}}}_i$ by initially sampling a noisy output $ \mathrm{\textbf{z}}_T$  and then gradually producing less-noisy samples $\mathrm{\textbf{z}}_{T-1}, \mathrm{\textbf{z}}_{T-2}, \dots, \mathrm{\textbf{z}}_0$. The diffusion model is trained with the following loss objective and once it has trained, the sampled latents represent $\hat{P}(\mathrm{z} \mid \mathrm{X}_{\text{target}})$:
\begin{equation} \label{eq:total_loss}
    \mathcal{L}_{\text{total}} = \lambda_1 \mathcal{L}_{\text{MSE}} +  \mathcal{L}_{\text{div}} + \lambda_2 \mathcal{L}_{\text{KL}} + \lambda_3 \mathcal{L}_g + \mathcal{L}_{\text{percep}}. 
\end{equation}

Each of the terms in the equation above are explained in more detail in this section. $\mathcal{L}_{\text{MSE}}$ is the standard MSE loss between the input latents,  $\mathrm{\textbf{z}}_i$ and the denoised output from the diffusion model, $\hat{\mathrm{\textbf{z}}}_i$. We find through supplementary experiments that optimizing with only an MSE objective is not enough to promote target sample diversity. Specifically, the inversion module shows excellent reconstruction ability even for out-of-distribution images, which lends to the images obtained from $\mathrm{\textbf{z}}_i$, ($ G_s(\mathrm{\textbf{z}}_i$)), to be a close reconstruction of the original $\mathrm{X}_i$. Thus, it follows that $\hat{P}(\mathrm{z}\mid \mathrm{X}_{\text{target}})$ is a close approximation of $P(\mathrm{z}\mid \mathrm{X}_{\text{target}})$. However, due to the limited target data, it can be difficult to ensure sample diversity through a simple MSE optimization. Instead, we leverage this strong advantage that inversion provides and employ some clever data augmentation approaches \cite{chong2022jojogan, thopalli2023sista}. 

A major advantage of our proposed method is that the latent sampler is trained in the $18\times512$ $\mathcal{W}^+$ space which is significantly lower-dimension than the original $1024\times1024$ image space the StyelGAN2 generator, $G_s$ is trained in. 
After obtaining $\hat{\mathrm{\textbf{z}}}_i$ from the diffusion model, we sample a random latent $\mathrm{\textbf{w}}$ from the pre-trained generator ($G_s$) latent space. It has been demonstrated that the StyleGAN2 latent space is disentangled \cite{wu2021stylespace}, and in the $18\times512$ latent vector, the style information is contained in the later layers while the semantic information is captured in first few layers of the $18 \times 512$ latent code. Abdal {\em et. al.} \cite{abdal2020image2stylegan++} successfully manipulate the style content in images using this assumption, and  one-shot style adaptation has also been achieved by leveraging data augmentation in addition to this knowledge \cite{chong2022jojogan}.  Leveraging the same idea, we modify $\hat{\mathrm{\textbf{z}}}_i$ to $\Bar{{\mathrm{\textbf{z}}}_i} = \alpha \hat{\mathrm{\textbf{z}}}_i + (1 - \alpha) \mathrm{\textbf{w}}$ where $\alpha$ controls the contribution of the randomly sampled latent $\mathrm{\textbf{w}}$. The modification, $\alpha\hat{\mathrm{\textbf{z}}}_i$ is only made to the latter style layers of $\hat{\mathrm{\textbf{z}}}_i$ since we want to produce diverse style variations of the denoised latent, $\hat{\mathrm{\textbf{z}}}_i$. We calculate $\mathcal{L}_{\text{div}}$ as $\|\mathrm{\textbf{z}}_i - \Bar{\mathrm{\textbf{z}}}_i\|^2_2$ and $\mathcal{L}_{\text{KL}}$ as the standard KL-divergence loss between $\Bar{\mathrm{\textbf{z}}}_i$ and $\mathrm{\textbf{z}}_i$. These losses combined push the model to focus on creating more diversity in the output samples. 

Preserving the geometry implies that distances, angles, and curvatures remain similar in both the low dimensional latent-space and high dimensional image spaces. To this end, we do not assume a flat geometry in the latent space \cite{liang2023learningposeimagemanifolds}. Further, preserving the pairwise dissimilarity in the tangent-space orientations requires the augmentation of the training set by calculating the tangent vectors. Thus to account for $\mathcal{L}_{\text{g}}$, we augment the denoised latent vector $\hat{\mathrm{\textbf{z}}}_i$ obtained from the latent sampler, and the corresponding image $\hat{\mathrm{X}}_i$ (where $\hat{\mathrm{X}}_i = G_s(\hat{\mathrm{\textbf{z}}}_i)$ and $G_s$ is kept frozen) to include tangent planes $T_{\mathrm{z}}$ and $T_{\mathrm{X}}$, respectively.   First, we find the pairwise L2 distance $d_{\mathrm{X}_i}$ between $\mathrm{X}_i$ and $\hat{\mathrm{X}}_i$ . We define the tangent vector $\Delta_{\mathrm{X}}$ as:
\begin{equation}
\label{eq:tangent_vector}
    \Delta_{\mathrm{X}} = (\mathrm{X}_i - \hat{\mathrm{X}}_i) / d_{\mathrm{X}_i}.
\end{equation}

To learn and estimate the tangent space of the generated images, we select the $k$ dominant singular value vectors of $\Delta_{\mathrm{X}}$ to obtain $T_\mathrm{x}$. We find that a higher value of $k$ reports better metrics for adaptation (more explained in section \ref{latent_sp_dim}). We repeat the steps above to find $T_{\mathrm{z}}$ using $\mathrm{\textbf{z}}_i$ and $\hat{\mathrm{\textbf{z}}}_i$. The pairwise distance metric $\mathcal{D}^{\mathrm{z,x}}$ is defined as:
\begin{equation}
    \mathcal{D}^{\mathrm{z,x}} = d(\left\| T_\mathrm{z} \right\|_\mathcal{F}, \left\| T_\mathrm{x} \right\|_\mathcal{F}),
\end{equation}
where, $\|\cdot\|_\mathcal{F}$ is the Frobenius norm and $d$ is any choice of distance measure (we choose cosine similarity). At this stage, we revisit our assumption that geometric elements (such as angles and curvatures) remain similar in both spaces. $\mathcal{D}^{\mathrm{z}, \mathrm{x}}$ is a measure of dissimilarity between the tangent planes in the image and latent space. Ideally, they would not be dissimilar, and thus,  we define $\mathcal{L}_g$ as: 
\begin{equation}
    \mathcal{L}_g = L_{MSE}(\mathcal{D}^{\mathrm{z,x}}, \mathds{1}).
\end{equation}
where $L_{MSE}$ is a MSE-loss calculated between the actual distance and a vector of 1's (ideal case). 

Finally, to ensure perceptual quality of the generated images from the denoised latents $\hat{\mathrm{\textbf{z}}}_i$, we leverage the VGG-16 perceptual loss, $\mathcal{L}_{\text{percep}}$ \cite{simonyan2014very}. For a batch of input latents to the diffusion model of size $B \times 18 \times 512$, $\mathrm{\textbf{z}}_i$ and the corresponding denoised output $\hat{\mathrm{\textbf{z}}}_i$ we use the pre-trained StyleGAN generator, $G_s$ to obtain $\mathrm{X}_i$ and $\hat{\mathrm{X}}_i$, respectively. Both $\mathrm{X}_i$ and $\hat{\mathrm{X}}_i$ are images and have shape $B \times 3 \times 1024 \times 1024$ and each batch of images contains $b$ elements. The VGG perceptual loss, $\mathcal{L}_{\text{percep}}^{\phi,b}$ for every $b^{th}$ image in the $i^{th}$ batch is defined as:
\begin{equation}
    \mathcal{L}_{\text{percep}}^{\phi,b}=\frac{1}{C_b H_b W_b}\left\|\phi_b(\hat{\mathrm{X}}_b)-\phi_b(\mathrm{X}_b)\right\|_2^2,
\end{equation}
where, $\phi$ represents pre-trained VGG feature extractors and $C_b, H_b, W_b$ denote the channel, height, and width of the $b^{th}$ image in the $i^{th}$ batch of size $B$. It is important to note that the VGG feature extractor is used to only extract the features, no update is made to $G_s$ as it is still frozen in this stage. The calculated loss $\mathcal{L}_{\text{percep}}$ is added to the total loss represented in \ref{eq:total_loss} and used to optimize the latent diffusion model and \textit {not the source generator}, $G_s$. In Equation \eqref{eq:total_loss}, $\lambda_1, \lambda_2$, and $\lambda_3$ denote tuning parameters that dictate the contribution of $\mathcal{L}_{\text{MSE}}$, $\mathcal{L}_{\text{KL}}$, and $\mathcal{L}_g$. Empirically, we set $\lambda_1, \lambda_2$, and $\lambda_3$ as 0.5, 0.1, and 0.2, respectively. 

\begin{figure*}
    \centering
    \includegraphics[width=\textwidth]{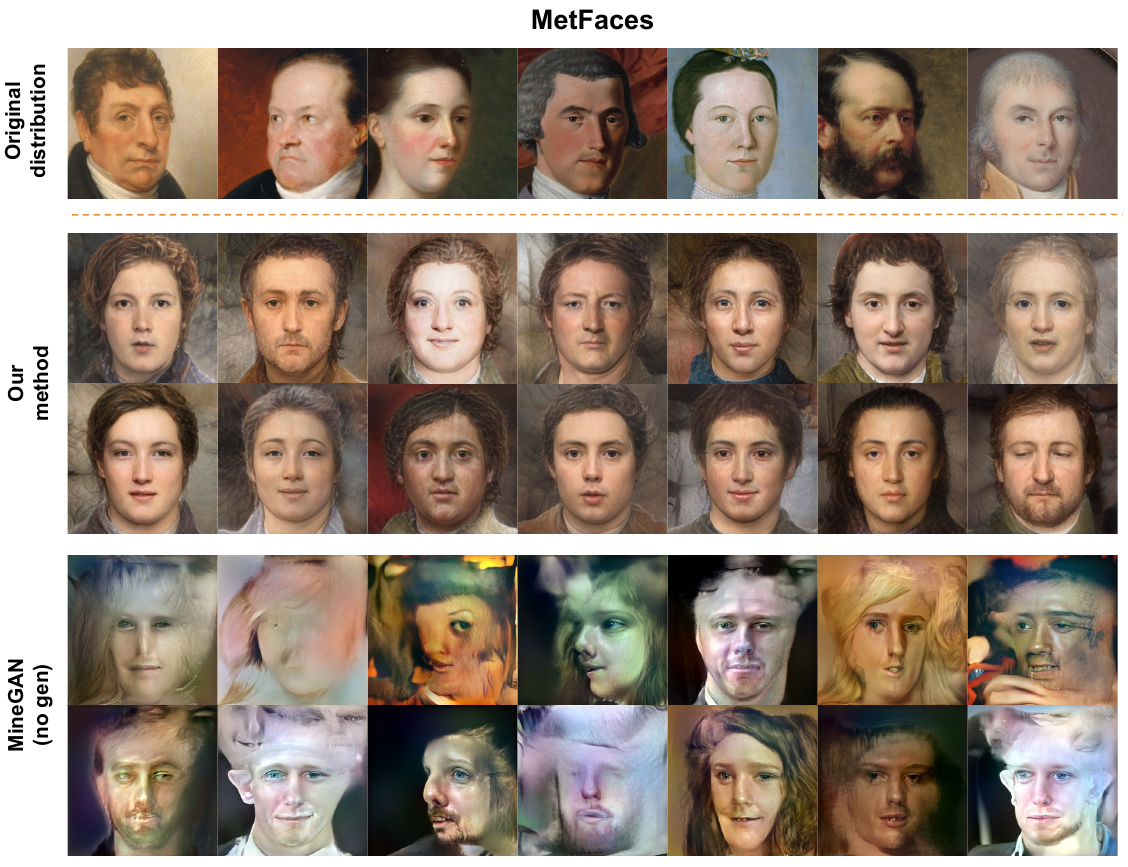}
    \caption{Adaptation performance of our pipeline for MetFaces using the pSp inverter. Our pipeline produces good quality, diverse images whereas the MineGAN baseline without generator update fails at the adaptation task. This demonstrates that learning a target distribution without updating the source generator is not a trivial task.}
    \label{fig:all_mets_adapt}
\end{figure*}

\section{Experiments}
\label{sec: exp}

\begin{figure*}
    \centering
    \includegraphics[width=\textwidth]{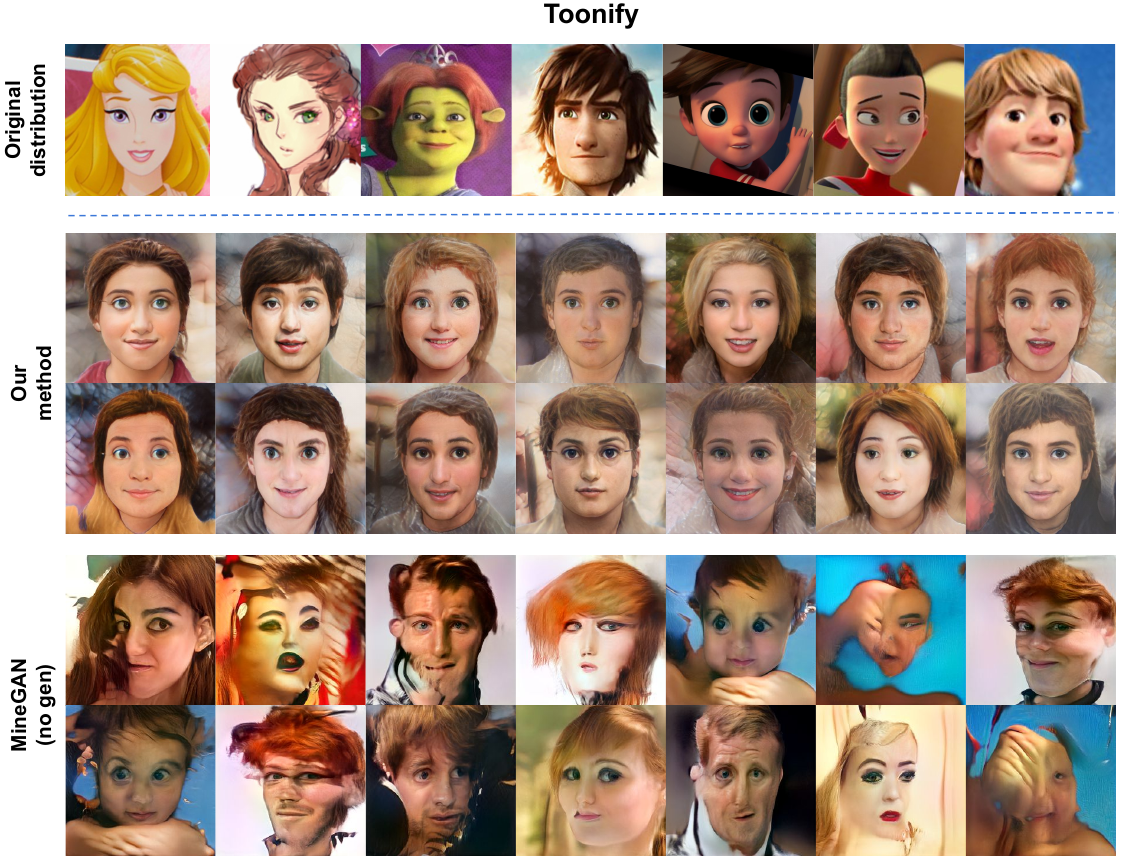}
    \caption{Adaptation performance of our pipeline for  Toonify using the pSp inverter. Following the trend observed in \ref{fig:all_mets_adapt} our pipeline shows excellent performance compared to the MineGAN implementation without the generator.}
    \label{fig:all_toon_adapt}
\end{figure*}

\subsection{Sampling from OOD target distribution}

To understand the adaptation ability of our pipeline, we use StyleGAN2 trained on FFHQ faces \cite{karras2019style} as our base source generator $G_s$,  and evaluate its performance on two out-of-domain (OOD) datasets: MetFaces and Toonify. The MetFaces dataset \cite{karras2020traininggenerativeadversarialnetworks} consists of 1336 high-quality PNG images at 1024×1024 resolution. This dataset is considered ``near OOD" for our experimental setup. On the other hand, the Toonify dataset \cite{pinkney2020resolution} consists of 317 cartoon images at $1024\times1024$ resolution. This OOD dataset is important for demonstrating the adaptation capabilities of our method in the case of limited target data. Visual results are shown in Figure \ref{fig:all_mets_adapt} for the MetFaces dataset and in Figure \ref{fig:all_toon_adapt} for the Toonify dataset.

\noindent{\textbf{Metrics and baselines.}} The Frechet Inception Distance (FID) \cite{heusel2017gans} and Inception Score (IS) \cite{salimans2016improved} are two widely used evaluation metrics to determine the diversity and fidelity of generative models. However, since FID is known to be unreliable for small datasets, we also consider the evaluation criteria presented in \cite{naeem2020reliable}. Traditional precision and recall metrics for GANs \cite{kynkaanniemi2019improved} suffer from the presence of outliers and computational inefficiency. The two new metrics -- density and coverage \cite{naeem2020reliable}, address these concerns by improving upon existing metric definitions. Density rewards samples in regions where real samples are densely packed while coverage measures the fraction of real samples whose neighborhood contains at least one fake sample. We implement our pipeline using the pSp and e4e inversion modules. We first create a test set by setting aside 836 images from MetFaces and 100 images from Toonify. We then invert the remaining images ($N_{target}$) using the chosen inverter to obtain the set of corresponding latent codes which are input to the diffusion model.

\begin{table*}[htb!]
\centering
\renewcommand{\tabcolsep}{1.0mm} 
\caption{The performance of our pipeline compared to two baselines: MineGAN without the generator update, and our method optimized with only an MSE objective ($\mathcal{L}_{\text{MSE}}$), with 500 MetFaces and 217 Toonify target images for the pSp inverter. The introduction of the new loss function in our method provides a boost in performance compared to $\mathcal{L}_{\text{MSE}}$. Additionally the MineGAN baseline shows severely decreased performance indicating that a simple auxillary MLP without generator fine-tuning is insufficient to learn the target distribution.}

\begin{center}

\scalebox{0.7}{
\begin{tabular}{lc|cccccl}
\hline
                                   &                                                                      & \textbf{FID$\downarrow$} & \textbf{Precision$\uparrow$} & \textbf{Recall$\uparrow$} & \textbf{Density} $\uparrow$     & \textbf{Coverage$\uparrow$} & \textbf{IS$\uparrow$} \\ \hline
\multirow{3}{*}{\textbf{MetFaces}} & \begin{tabular}[c]{@{}c@{}}MineGAN++ \\ (w/o generator)\end{tabular} & 130.149                  & 0.674/0.167                  & 0.003/0.073               & 0.312/0.125          & 0.308/0.097                 & 1.517                 \\
                                   & $\mathcal{L}_{\text{MSE}}$                                                               & 1.835                    & 0.972/0.043                  & 0.007/0.025               & \textbf{1.198/0.256} & 0.508/0.088                 & 9.743                 \\
                                   & Our method                                                           & \textbf{0.925}           & \textbf{0.988/0.025}         & \textbf{0.014/0.033}      & 1.133/0.332          & \textbf{0.656/0.105}        & \textbf{11.651}       \\ \hline
\multirow{3}{*}{\textbf{Toonify}}  & \begin{tabular}[c]{@{}c@{}}MineGAN++ \\ (w/o generator)\end{tabular} & 201.623                  & 0.496/0.286                  & 0.004/0.237               & 0.523/0.313          & 0.385/0.289                 & 1.774                 \\
                                   & $\mathcal{L}_{\text{MSE}}$                                                               & 5.859                    & 0.917/0.289                  & 0.065/0.113               & 0.628/0.245          & 0.694/0.399                 & 6.247                 \\
                                   & Our method                                                           & \textbf{3.921}           & \textbf{0.932/0.289}         & \textbf{0.073/0.113}      & \textbf{1.994/0.399} & \textbf{0.792/0.294}        & \textbf{8.836}        \\ \hline
                                   
\end{tabular}
}

\label{tab:baselines}
\end{center}
\end{table*}

\begin{table}[ht!]
\renewcommand{\tabcolsep}{1.2mm}
\caption{Comparing our method to MineGAN \cite{wang2024minegan++} with generator update for MetFaces (N=500) and Toonify (N=217) images. Although generator manipulation gives better metrics in some instances, our pipeline approach these metrics closely. Note, MetF. denotes MetFaces.}
\begin{center}

\scalebox{0.9}{
\begin{tabular}{cc|ccccc}
\hline
                                   &                                                                       & \textbf{FID} $\downarrow$ & \textbf{Precision} $\uparrow$ & \textbf{Recall} $\uparrow$ & \textbf{Density} $\uparrow$ & \textbf{Coverage}$\uparrow$ \\ \hline
\multirow{2}{*}{\rotatebox[origin=c]{90}{\textbf{MetF.}}} & \textbf{\begin{tabular}[c]{@{}c@{}}MineGAN\\ (with gen)\end{tabular}} & 0.699        & 0.972/0.032        & 0.206/0.111     & 1.005/0.231      & 0.728/0.108       \\
                                   & \textbf{\begin{tabular}[c]{@{}c@{}}Our\\ method\end{tabular}}         & 0.925        & 0.988/0.025        & 0.014/0.033     & 1.133/0.332      & 0.656/0.105       \\ \hline
\multirow{2}{*}{\rotatebox[origin=c]{90}{\textbf{Toonify}}}  & \textbf{\begin{tabular}[c]{@{}c@{}}MineGAN\\ (with gen)\end{tabular}} & 2.425        & 0.969/0.108        & 0.750/0.177     & 1.073/0.311      & 0.979/0.072       \\
                                   & \textbf{\begin{tabular}[c]{@{}c@{}}Our\\ method\end{tabular}}         & 3.921        & 0.932/0.289        & 0.073/0.113     & 1.994/0.399      & 0.792/0.294       \\ \hline
\end{tabular}}

\label{tab:minegan_gen_comparison}

\end{center}
\end{table}


Since there is no direct comparison to our method, we consider two relevant baselines - first, a ``miner" based adaptation \cite{wang2024minegan++} where we train a simple MLP in the latent space that identifies which part of the generative distribution of a pre-trained GAN outputs samples closest to the target domain. Second, we modify Equation \eqref{eq:total_loss} to contain only  $\mathcal{L}_{\text{MSE}}$ ($\lambda_1 = 1$) and implement our method with this loss only. This serves as an important baseline to demonstrate the contribution of $\mathcal{L}_g$. As discussed in the prior sections, freezing the generator limits the adaptation ability and we find this reflected in the results in Table \ref{tab:baselines}. In Table \ref{tab:minegan_gen_comparison} we compare our method to the case where MineGAN \cite{wang2024minegan++} allows for generator update during the adaptation process.  In Table \ref{tab:adaptation_metrics}, we report evaluation metrics by varying the number of available target images ($N_{target}$) and then evaluating the results against the test set by sampling from the diffusion model after training.

\noindent{\textbf{Training details.}} For the adaptation experiments, the set of training latents is first normalized as $\mathrm{z} = (\mathrm{z} - \mu) / \gamma$ where $\mu$ and $\gamma$ represent the mean and standard deviation of the inverted latents. They are then passed to the diffusion sampler in batches of size $B$ with a learning rate of $1e^{-3}$ using the Adam optimizer \cite{kingma2014ga} and trained for $M$ epochs. For the MetFaces task, $B$ = 8 and $M$ = 1500. For Toonify, $B = 8$ for $N_{target} = 217$ , $B = 4$ for $N_{target} = 108 $ and $N_{\text{target}} = 54$, respectively, and $M = 1500$.

\begin{table}[htb!]
\centering
\renewcommand{\tabcolsep}{1.2mm}
\begin{center}
\caption{The performance on the learned target distribution using our pipeline across the two inversion modules - pSp and e4e. We vary the number of target images and observe the trends in adaptation performance. Our pipeline shows graceful degradation of performance as the target set reduces for the Toonify dataset.}
{
\begin{tabular}{cccccccc}
\hline
                                                        &                                                    & \textbf{$N_{target}$}             & \textbf{FID$\downarrow$} & \textbf{Precision$\uparrow$} & \textbf{Recall$\uparrow$} & \textbf{Density}$\uparrow$ & \textbf{Coverage$\uparrow$} \\ \hline
\multicolumn{1}{c|}{\multirow{6}{*}{\rotatebox[origin=c]{90}{\textbf{MetFaces}}}} & \multicolumn{1}{c|}{\multirow{3}{*}{\textbf{pSp}}} & \multicolumn{1}{c|}{500} & 0.925                  & 0.988/0.025                  & 0.014/0.033               & 1.133/0.332      & 0.606/0.105                 \\
\multicolumn{1}{c|}{}                                   & \multicolumn{1}{c|}{}                              & \multicolumn{1}{c|}{400} & 0.928                  & 0.987/0.013                  & 0.011/0.023               & 1.157/0.322      & 0.565/0.124                 \\
\multicolumn{1}{c|}{}                                   & \multicolumn{1}{c|}{}                              & \multicolumn{1}{c|}{300} & 0.936                  & 0.994/0.015                  & 0.006/0.018               & 1.138/0.231      & 0.452/0.079                 \\ \cline{2-8} 
\multicolumn{1}{c|}{}                                   & \multicolumn{1}{c|}{\multirow{3}{*}{\textbf{e4e}}} & \multicolumn{1}{c|}{500} & 1.110                  & 0.966/0.042                  & 0.018/0.032               & 0.924/0.349      & 0.451/0.136                 \\
\multicolumn{1}{c|}{}                                   & \multicolumn{1}{c|}{}                              & \multicolumn{1}{c|}{400} &    1.548        &  0.907/0.038    &   0.016/0.020     &  0.915/0.219                 &     0.517/0.128                         \\
\multicolumn{1}{c|}{}                                   & \multicolumn{1}{c|}{}                              & \multicolumn{1}{c|}{300} &      2.731   &    0.899/0.089         &     0.015/0.009                       &    0.830/0.306               &        0.451/0.120                      \\ \hline
\multicolumn{1}{c|}{\multirow{6}{*}{\rotatebox[origin=c]{90}{\textbf{Toonify}}}}  & \multicolumn{1}{c|}{\multirow{3}{*}{\textbf{pSp}}} & \multicolumn{1}{c|}{217} & 3.921                   & 0.932/0.289                  & 0.073/0.113               & 1.994/0.399      & 0.792/0.294                 \\
\multicolumn{1}{c|}{}                                   & \multicolumn{1}{c|}{}                              & \multicolumn{1}{c|}{108} & 4.010                   & 0.927/0.289                  & 0.042/0.081               & 1.025/0.442      & 0.771/0.286                 \\
\multicolumn{1}{c|}{}                                   & \multicolumn{1}{c|}{}                              & \multicolumn{1}{c|}{54}  & 4.396                   & 0.920/0.277                  & 0.030/0.083               & 1.090/0.543      & 0.640/0.315                 \\ \cline{2-8} 
\multicolumn{1}{c|}{}                                   & \multicolumn{1}{c|}{\multirow{3}{*}{\textbf{e4e}}} & \multicolumn{1}{c|}{217} &  6.776                         &              0.979/0.072                 &         0.029/0.045                   &       1.246/0.399            &          0.875/0.232                    \\
\multicolumn{1}{c|}{}                                   & \multicolumn{1}{c|}{}                              & \multicolumn{1}{c|}{108} &    7.410                       &         0.932/0.087      &    0.022/0.038   &     1.045/0.282              &        0.807/0.246                      \\
\multicolumn{1}{c|}{}                                   & \multicolumn{1}{c|}{}                              & \multicolumn{1}{c|}{54}  &  8.475                         &         0.879/0.057                      &       0.018/0.034                     &      1.172/0.358             &      0.698/0.164

                        \\ \hline
\end{tabular}
}

\label{tab:adaptation_metrics}

\end{center}
\end{table}

\subsection{Controlled attribute generation}

\begin{figure}[htb!]
    \centering
    \includegraphics[width=\textwidth]{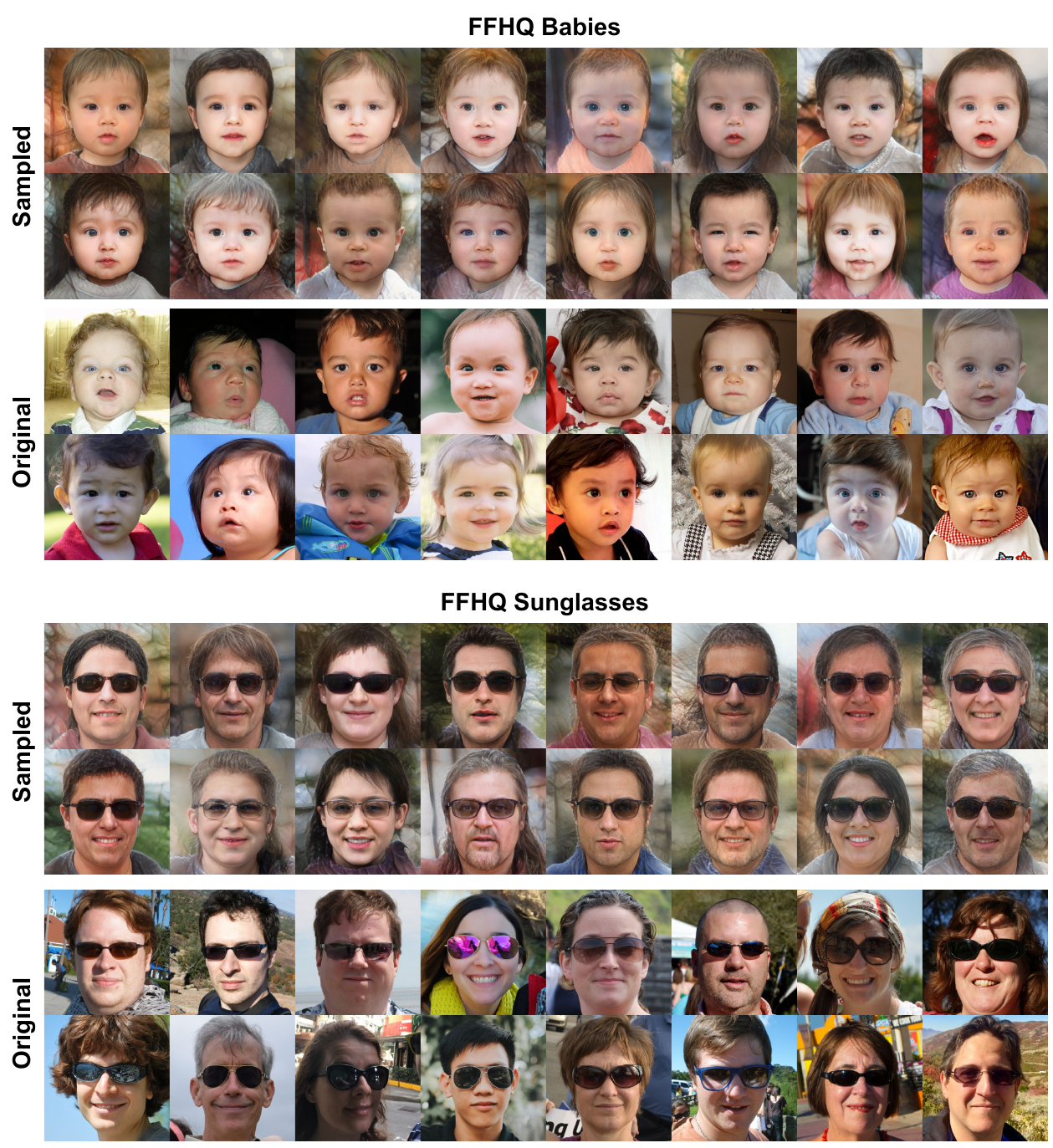}
    \caption{The performance of our pipepline for the ``babies" and ``sunglasses" attribute in the FFHQ dataset. The learned distribution (top row) closely resembles the original distribution (bottom row).}
    \label{fig:ffhq_attributes}
\end{figure}

In addition to adaptation, we investigate controlled image generation using our pipeline by emphasizing specific FFHQ image attributes. Since any face attribute is a subset of the original FFHQ image dataset that StyleGAN2 is originally trained on, this experiment demonstrates the ability of the diffusion model to learn specific distributions in the latent space, independent of any bias from the target set or the inverter in addition to providing the ability to unconditionally sample from a specific attribute. We construct target sets only from babies and faces with sunglasses and report the results in Table \ref{tab:attribute_metrics} using the pSp inverter. In Figure \ref{fig:ffhq_attributes}, we observe that our method produces high quality diverse samples from the target attribute distribution thus substantiating the claim that learning distributions in the latent space using our pipepline is data-efficient.

\begin{table}[ht!]
\begin{center}
\caption{Controlled generation with limited target data from (a) FFHQ babies and (b) FFHQ sunglasses. Even at a small number of target images ($N_{target}=10$ for (a) babies and (b) sunglasses) shows promising results. }
\begin{tabular}{cc|ccccc}
\hline
                                     & \textbf{$N_{target}$} & \textbf{FID}$\downarrow$& \textbf{Precision}$\uparrow$& \textbf{Recall}$\uparrow$& \textbf{Density}$\uparrow$& \textbf{Coverage}$\uparrow$\\ \hline
\multirow{3}{*}{\textbf{(a)}}     & \textbf{150} & 0.717& 0.992/0.016        & 0.038/0.038     & 1.880/0.398        & 0.876/0.090       \\
                                     & \textbf{80}  & 0.758& 0.995/0.012        & 0.031/0.046     & 2.042/0.416        & 0.856/0.090       \\
                                     & \textbf{10}  & 0.793& 0.999/0.006        & 0.015/0.021     & 2.351/0.420        & 0.871/0.085       \\ \hline
\multirow{3}{*}{\textbf{(b)}} & \textbf{150} & 4.938& 0.997/0.021        & 0.272/0.196     & 1.380/0.221        & 0.996/0.030       \\
                                     & \textbf{80}  & 5.404& 0.996/0.021        & 0.239/0.188     & 1.294/0.272        & 0.993/0.069       \\
                                     & \textbf{10}  & 5.849& 0.998/0.016        & 0.191/0.151     & 1.308/0.210        & 0.971/0.021       \\ \hline
\end{tabular}

\label{tab:attribute_metrics}
\end{center}
\end{table}

\subsection{Text based manipulation}

\begin{figure}[ht!]
    \centering
    \includegraphics[width=1.0\columnwidth]{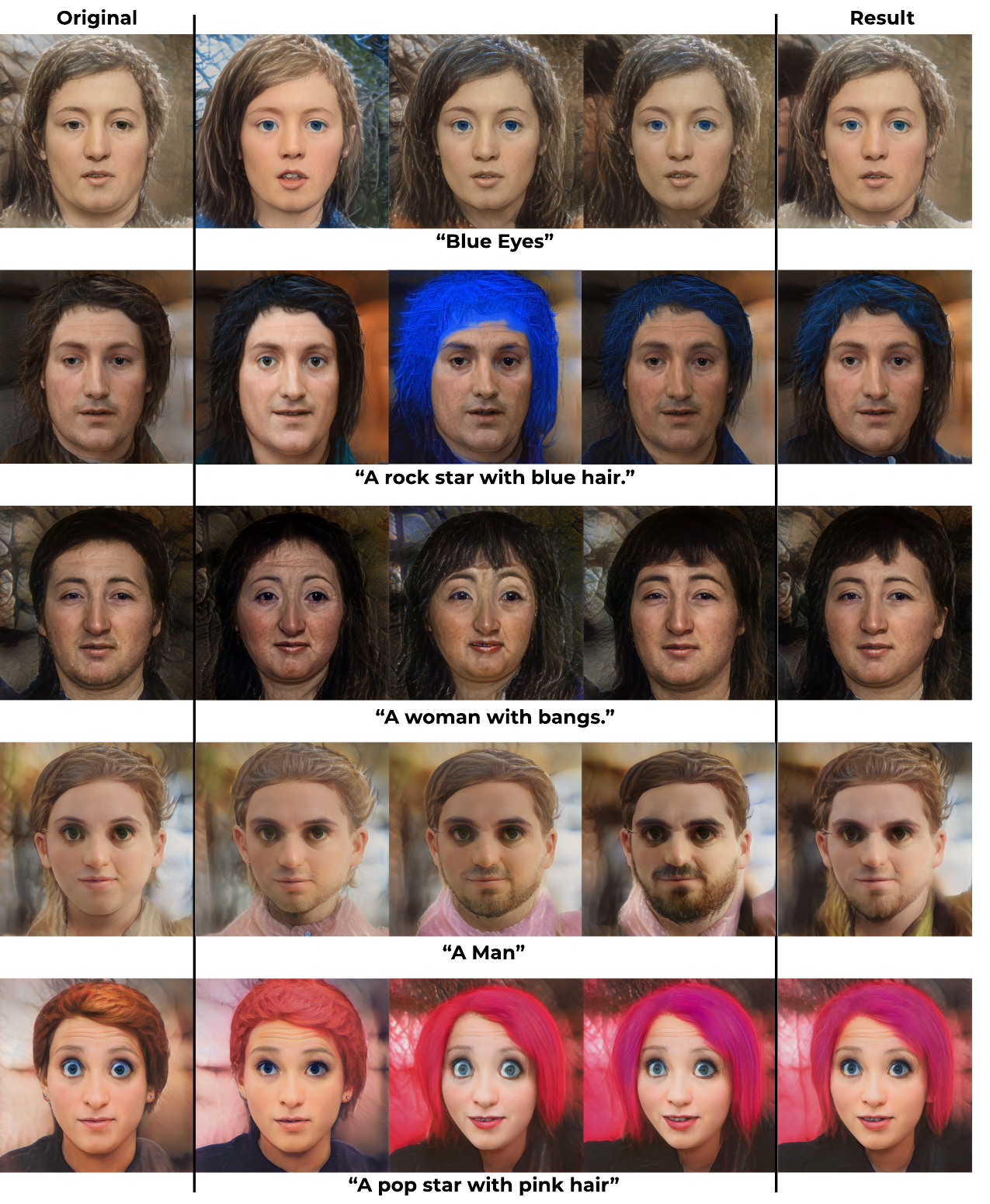}
    \caption{Adaptation by text prompt using StyleCLIP. We visualize the ``walk" from the adapted latents obtained from the latent sampler and the final CLIP modified results. The first three rows feature images obtained from latents adapted to MetFaces and the last two rows feature the same for Toonify. The images in the middle columns represent different steps in the StyleCLIP adaptation process and were selected arbitrarily during training. }
    \label{fig:StyleCLIPFigure}
\end{figure}

We conduct an auxiliary experiment to demonstrate our pipeline's compatibility for CLIP-based editing should a specific application require it. To this end, the latent sampler is used to sample from the learned target distribution and obtain target latent vector $\mathrm{\textbf{z}}_s$. Following StyleCLIP \cite{patashnik2021styleclip}, we choose an arbitrary text prompt $t$ and introduce the pair $(\mathrm{\textbf{z}}_s, t)$ to the StyleCLIP encoder $G_{\text{CLIP}}$. The initial latent code, $\mathrm{\textbf{z}}_s$ is optimized to produce $\hat{\mathrm{\textbf{z}}}_s$ in order to match the text prompt $t$  using the CLIP-based loss function $\mathcal{L}_{\text{CLIP}}(\cdot)$. For every training iteration, $\mathcal{L}_{\text{CLIP}}(\cdot)$ minimizes the cosine similarity distance between $\hat{\mathrm{X}}_s = G_{\text{CLIP}} (\hat{\mathrm{\textbf{z}}}_s)$  and $t$ in the CLIP latent space according to the following loss function: 
\begin{equation} \label{eq:clip_loss}
    \mathcal{L}(\mathrm{\textbf{z}}) = \mathcal{L}_{\text{CLIP}}(\hat{\mathrm{X}}_s, t) + \lambda_{\mathcal{L}\text{2}} \left\| \mathrm{\textbf{z}}_s - \hat{\mathrm{\textbf{z}}}_s\right\|_{\text{2}}.
\end{equation}

$G_{\text{CLIP}}$ is kept frozen and only the initial latent code $\mathrm{\textbf{z}}_s$ is optimized using an Adam optimizer with a learning rate = 0.1. The weighted parameter $\lambda_{\mathcal{L}\text{2}}$ controls the MSE loss between the $\mathrm{\textbf{z}}_s$ and $\hat{\mathrm{\textbf{z}}}_s$ and we set it to 0.005. As shown in Figure \ref{fig:StyleCLIPFigure}, the adapted latents obtained from our pipeline can be further modified through text prompts. 
This is a significant observation since now we not only have the ability to unconditionally sample from a target distribution without manipulating the source generator, but also the ability to edit the target images according to a specific application. 

\subsection{Ablations}

\paragraph{Contribution of loss terms.}

\begin{figure}[ht!]
    \centering
    \includegraphics[width=0.85\columnwidth]{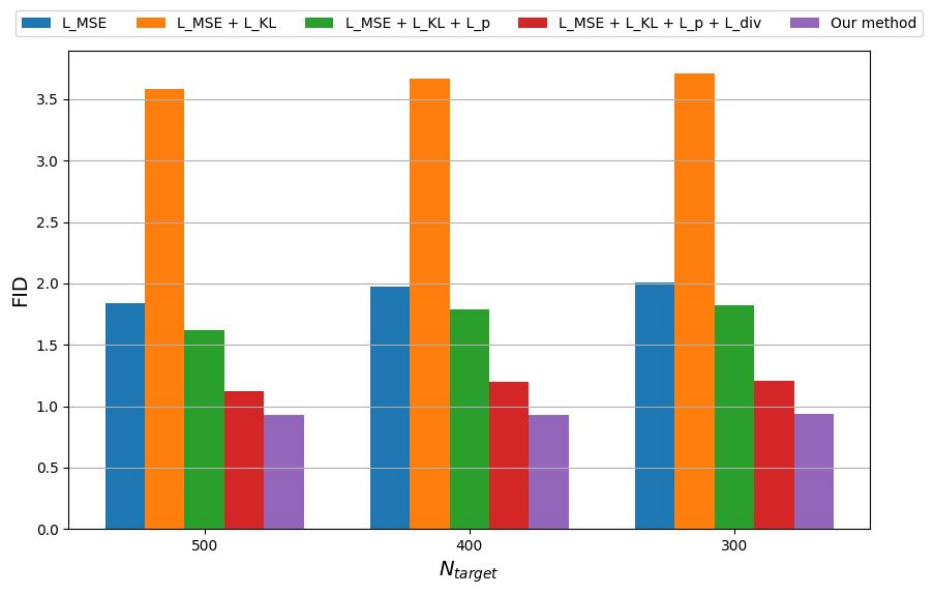}
    \caption{FID scores on varying $N_{target}$ images from MetFaces for the pSp inverter. Progressively adding loss terms to Equation \eqref{eq:total_loss} displays that adding $\mathcal{L}_g$ gives the best FID score. $L_{MSE}, L_{KL}, L_p,L_{div}$ denote $\mathcal{L}_{\text{MSE}}, \mathcal{L}_{\text{KL}}, \mathcal{L}_{\text{percep}}, \mathcal{L}_{\text{div}}$ from Equation \eqref{eq:total_loss} respectively. }
    \label{fig:baselines_vs_fid}
\end{figure}

\begin{figure}[htb!]
    \centering
    \includegraphics[width=1\linewidth]{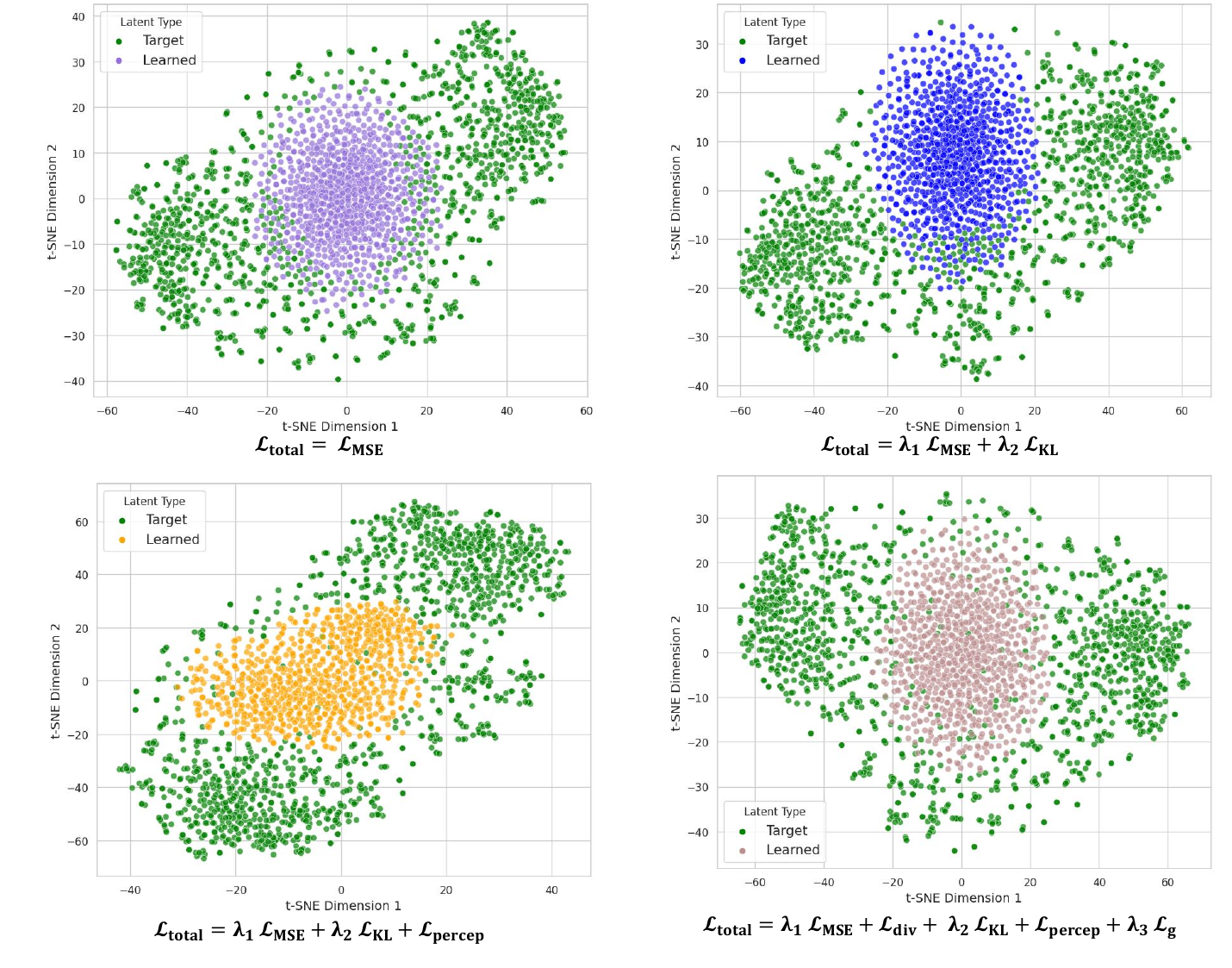}
    \caption{Analysis on loss functions in Equation \eqref{eq:total_loss} to investigate the contribution of each loss term. In each subplot, the sampled latents from the learned distribution are represented in various colors while the original target latents are displayed in green.}
    \label{fig:tsne_loss}
\end{figure}

We measure clustering performance using t-SNE \cite{vandermaaten08a} and conduct a study on progressively adding loss terms in Equation \eqref{eq:total_loss}. In Figure \ref{fig:baselines_vs_fid}, we report the adaptation performance (through FID score) while varying number of available target images from MetFaces.  We notice that adding the KL-divergence loss term, while slightly increasing overall diversity, creates a significant drop in the overall model performance. However, progressively adding more loss terms and finally adding $\mathcal{L}_g$ pushes our pipeline to report the best performance. Figure \ref{fig:tsne_loss} analyzes the effect of each loss term in context of the latents sampled during inference.  We first train the latent model with $\mathcal{L}_{\text{MSE}}$ only with $\lambda_1 = 1$ (top left graph with purple points) and visualize the results. This is one of our baselines for evaluating the performance of the adaptation pipeline. In conjunction with the previously reported metrics, from this t-SNE plot it is clear that the learned distribution does not capture the target distribution. Adding $\mathcal{L}_{\text{KL}}$ to this gives an interesting insight in the top-right t-SNE plot (blue points). Although the spread of the cluster denotes more diversity, the learned distribution resides beyond the bounds of the original inverted latents. For this realization  the latent model is trained with $\lambda_1 \mathcal{L}_{\text{MSE}} + \lambda_2 \mathcal{L}_{\text{KL}}$ where $\lambda_1 =  0.5 $ and $\lambda_2 = 0.1$.
Introducing $\mathcal{L}_{\text{percep}}$ (yellow points) slightly improves the performance from the previous case. And finally, we visualize the t-SNE embeddings after training with the final loss function in Equation \eqref{eq:total_loss} (pink points).  

\paragraph{Adaptation for limited target data.}

\begin{figure}[htb!]
    \centering
    \includegraphics[width=\columnwidth]{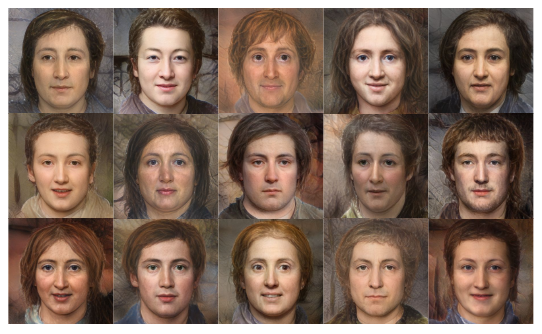}
    \caption{Output from our pipeline trained with 50 MetFaces target samples. Visually, we observe a variety of poses, expression, hair colour, gender, and overall art style. These images reflect the metrics reported in Table \ref{tab:lg_vary_n}.}
    \label{fig: mets_50_lg}
    
\end{figure}

As discussed previously, it is difficult to achieve good adaptation metrics in a setting where generator update is not possible and there is limited target data. We theorize that the introduction of tangent planes and the assumptions regarding the dimensionality of the latent space gives key insight into adapting generative models in this setting. To this end, we study the effectiveness of $\mathcal{L}_g$ in achieving adaptation for a limited number of target images using the pSp inverter for the MetFaces dataset. We report the metrics in Table \ref{tab:lg_vary_n} and a small sample of the generated images (Figure \ref{fig: mets_50_lg}).

\begin{table}[htb!]
\begin{center}
 \caption{We train the latent sampler with small number of target MetFaces images to demonstrate our pipeline's ability to perform adaptation in settings with limited target data. The metrics above are reported for latents obtained from pSp and by setting $\lambda_1 = 0.5, \lambda_2 = 0.1, \lambda_3 = 0.2$ in Equation \eqref{eq:total_loss}.}
\begin{tabular}{c|ccccc}
\hline
             \textbf{$N_{target}$} & \textbf{FID}$\downarrow$& \textbf{Precision}$\uparrow$& \textbf{Recall}$\uparrow$& \textbf{Diversity}$\uparrow$& \textbf{Coverage}$\uparrow$\\ \hline
\textbf{100} & 0.918& 0.992/0.021        & 0.004/0.010     & 1.146/0.333        & 0.542/0.105       \\
\textbf{50}  & 0.931& 0.891/0.006        & 0.003/0.006     & 1.320/0.392        & 0.519/0.131       \\
\textbf{10}  & 0.954& 0.889/0.020        & 0.003/0.010     & 1.216/0.332        & 0.493/0.093    \\ \hline  
\end{tabular}

\label{tab:lg_vary_n}
\end{center}
\end{table}

\paragraph{Latent space dimensionality.} \label{latent_sp_dim}

In Table \ref{tab:dim-performance}, we vary the number of singular values used in Equation\eqref{eq:tangent_vector} and report metrics on the adaptation performance for the MetFaces dataset. We find that assuming $k=10$ gives us the best performance for the pSp inverter.

\begin{table}[ht!]
\caption{Performances on assumed dimensionality of the latent space (in rows). Assuming the latent space manifold is low dimension shows overall decreased adaptation metrics.}
\begin{center}

\begin{tabular}{c|ccccc}

\hline
\textbf{$k$}   & \textbf{FID}$\downarrow$& \textbf{Precision}$\uparrow$& \textbf{Recall}$\uparrow$& \textbf{Density}$\uparrow$& \textbf{Coverage}$\uparrow$\\ \hline
\textbf{3}  & 0.931& 0.995/0.019        & 0.007/0.016     & 1.230/0.374      & 0.429/0.114       \\ \hline
\textbf{5}  & 0.925& 0.988/0.014        & 0.009/0.024     & 1.121/0.291      & 0.590/0.068       \\ \hline
\textbf{10} & 0.924& 0.988/0.025& 0.014/0.033& 1.133/0.332      & 0.606/0.105\\ \hline
\end{tabular}

\label{tab:dim-performance}
\end{center}
\end{table}



\section{Discussion and conclusion}

In this paper, we present a novel approach to GAN adaptation without requiring the manipulation or retraining of the source generator. By preserving pair-wise geometric distances and leveraging the advantage of state-of-the-art inverters, our pipeline is able to successfully sample from a target distribution. An interesting observation that we make is a relatively light-weight latent generative model is sufficient to achieve this, however, to ensure diversity of generated samples, simply introducing traditional diversity boosting losses (such as KL-divergence) are insufficient. Such an approach leads to slightly increased diversity but fails to learn the target distribution effectively. Thus, introducing a geometric prior benefits the adaptation performance and boosts diversity of sampled images for real distribution shifts. Additionally, we emphasize the main advantage of our pipeline - achieving adaptation in settings where access to the source generator may not be possible. With recent focus on the ethical implications of AI in general, and data ownership in particular, our approach presents an option for end-users to perform adaptation of generative models without having to transfer their own data (e.g. images) to commercial entities.


The performance of our pipeline suffers in cases where the inverters cannot project far-OOD images onto the latent manifold of the source generator. For instance, when inverting Ukiyoe images \cite{pinkney2020ukiyoe} we find that even state of the art inverters (such as ReStyle\cite{alaluf2021restyle}) are not able to correctly reconstruct the target images. 
We leave this implementation as future work. In conclusion, the current pipeline shows excellent promise for the adaptation of blackbox generative models in application-specific, data-constrained domain adaptation tasks.

\section*{Acknowledgements}
This work was supported by NSF grant 2323086. This work was supported in part by Seoul National University of Science and Technology.

\bibliography{references}

\end{document}